\documentclass{article}
\usepackage{ijcai19}

\usepackage{times}
\usepackage{soul}
\usepackage{url}
\usepackage[hidelinks]{hyperref}
\usepackage[utf8]{inputenc}
\usepackage[small]{caption}
\usepackage{graphicx}
\usepackage{amsmath}
\usepackage{booktabs}
\usepackage{xcolor}

\usepackage{caption} 
\usepackage{amsfonts,amssymb}
\usepackage{threeparttable}
\usepackage{booktabs}
\usepackage{diagbox}
\usepackage{colortbl}

\usepackage{multirow}

\usepackage{amsmath,bm}
\usepackage{dsfont}
\usepackage{color} 
\usepackage[switch]{lineno}  %

\title{AutoAlign: Pixel-Instance Feature Aggregation \\for Multi-Modal 3D Object Detection}

\author{
	Zehui Chen\textsuperscript{\rm 1}, 
	Zhenyu Li\textsuperscript{\rm 2}, 
	Shiquan Zhang\textsuperscript{\rm 3}, 
	Liangji Fang\textsuperscript{\rm 3}
	Qinghong Jiang\textsuperscript{\rm 3}, \\
	Feng Zhao*\textsuperscript{\rm 1},
	Bolei Zhou\textsuperscript{\rm 4}, 
	Hang Zhao\textsuperscript{\rm 5}
\affiliations
	\textsuperscript{\rm 1} University of Science and Technology,
	\textsuperscript{\rm 2} Harbin Institute of Technology \\
	\textsuperscript{\rm 3} SenseTime Research,
	\textsuperscript{\rm 4} The Chinese University of Hong Kong,
	\textsuperscript{\rm 5} IIIS, Tsinghua University\\
}

\begin{document}

\maketitle

\begin{abstract}

Object detection through either RGB images or the LiDAR point clouds has been extensively explored in autonomous driving. However, it remains challenging to make these two data sources complementary and beneficial to each other.  
In this paper, we propose \textit{AutoAlign}, an automatic feature fusion strategy for 3D object detection. Instead of establishing deterministic correspondence with camera projection matrix, we model the mapping relationship between the image and point clouds with a learnable alignment map. This map enables our model to automate the alignment of non-homogenous features in a dynamic and data-driven manner. 
Specifically, a cross-attention feature alignment module is devised to adaptively aggregate \textit{pixel-level} image features for each voxel. To enhance the semantic consistency during feature alignment, we also design a self-supervised cross-modal feature interaction module, through which the model can learn feature aggregation with \textit{instance-level} feature guidance. 
Extensive experimental results show that our approach can lead to 2.3 mAP and 7.0 mAP improvements on the KITTI and nuScenes datasets, respectively. Notably, our best model reaches 70.9 NDS on the nuScenes testing leaderboard, achieving competitive performance among various state-of-the-arts.  


\end{abstract}
\section{Introduction}
Recent advances in deep learning bring rapid progress in autonomous driving. 3D object detection through LiDAR points plays an essential role in understanding the surroundings of the vehicle. LiDAR points can capture precise 3D spatial information for object detection, however, they often suffer from both the lack of semantic information and sparsity of reflected points, leading to failures under foggy or crowded circumstances. Compared to point clouds, RGB images have better strength in providing semantic and long-distance information. Hence, many approaches explore the data fusion of RGB camera and LiDAR sensors to improve the performance of 3D object detection. 


\begin{figure}[ht!]
   \includegraphics[width=0.47\textwidth]{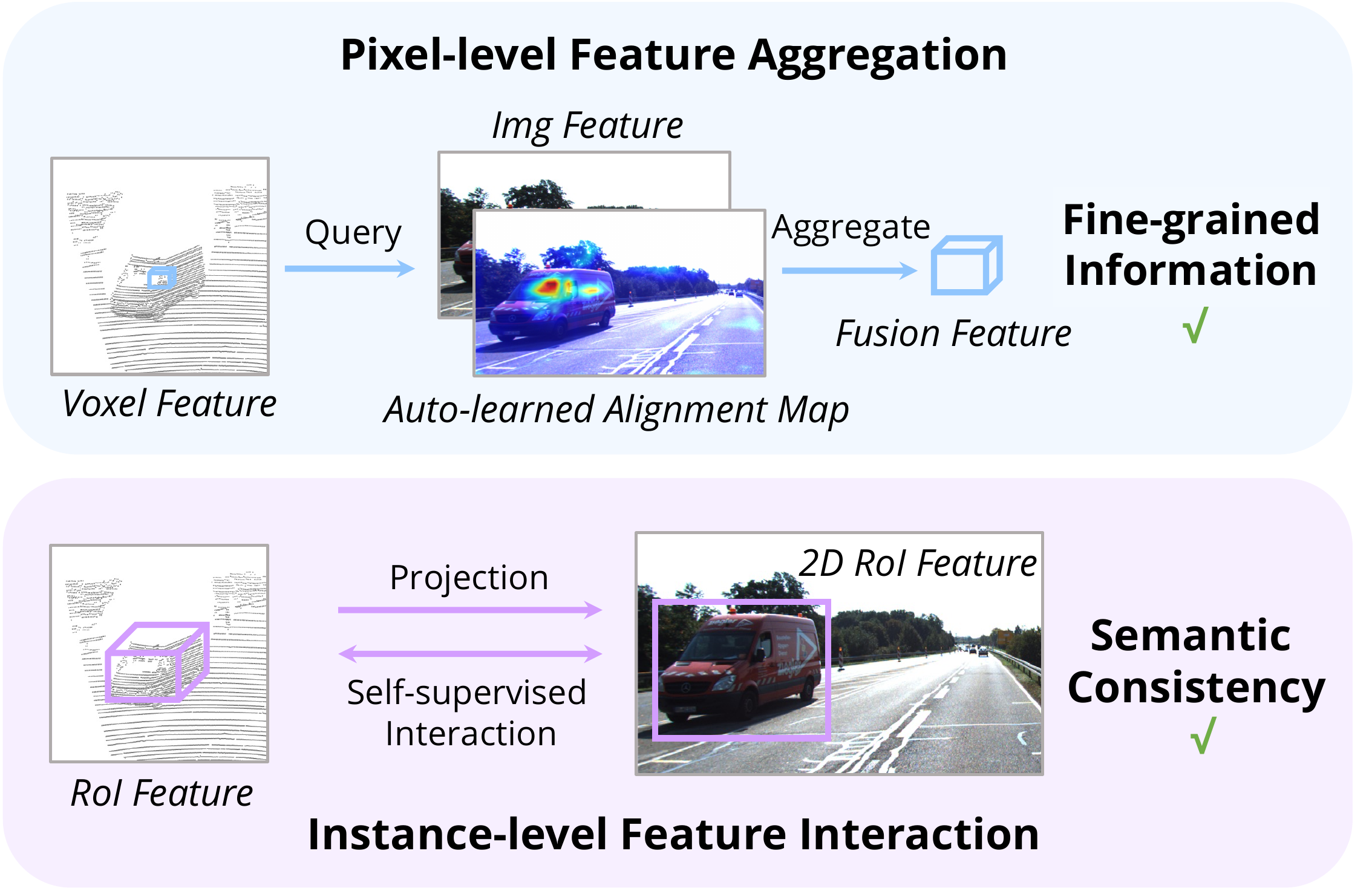}
\caption{The illustration of interactions between image and point cloud in AutoAlign. The interactions act on two levels: (i) the pixel-level feature aggregation preserves the fine-grained RGB features from images and (ii) the instance-level feature interaction enhances the semantic consistency between non-homogenous representations. 
}
\vskip-1em
\label{fig:intro_pdf}
\end{figure}

Multi-modal 3D object detectors can be roughly grouped into two categories: decision-level fusion and feature-level fusion. The former detects objects in respective modalities and then ensembles the boxes together in the 3D space~\cite{pang2020clocs}. Different from the decision-level fusion, the feature-level fusion combines the multi-modal features into a single representation from which the objects are detected. Therefore, the detector can fully utilize features from different modalities at the inference stage. In light of this, more approaches for feature-level fusion have been recently developed. One line of work~\cite{vora2020pointpainting,xie2020pi} projects each point to the image plane and gets the corresponding image features through bilinear interpolation. Although the feature aggregation is performed delicately at the \textit{pixel level}, in doing so we will lose the dense patterns in the image domain due to the sparsity of the fusion points, \textit{i.e.,} breaking the \textit{semantic consistency} in the image features. Another line of work~\cite{chen2017multi} uses initial proposals provided by 3D detectors to obtain respective RoI features at different modalities and concatenates them together for feature fusion. It maintains the semantic consistency by conducting \textit{instance-level} fusion, however, it suffers from the coarse feature aggregation and the absence of 2D information at the initial proposal generation phase. 

To take the best of these two types of approaches, we propose an integrated multi-modal feature fusion framework for 3D object detection, named \textbf{AutoAlign}. It enables the detector to aggregate cross-modal features in an adaptive way, which proves to be effective in modeling relationships between non-homogenous representations. Meanwhile, it takes advantage of the fine-grained feature aggregation at the pixel level, but at the same time preserves the semantic consistency through instance-wise feature interaction (see Figure \ref{fig:intro_pdf}).

Specifically, to preserve concrete details in the RGB data, we design a Cross-Attention Feature Alignment (CAFA) module, which dynamically attends the pixel-level features from the image and maintains the efficiency by fusing features at a higher 3D level (pillar or voxel). Each voxel feature will query the whole image plane to get a pixel-wise semantic alignment map. Then, CAFA aggregates the image features based on the alignment map and concatenates them together with the original 3D features. In order to facilitate the learning of semantic consistency between point clouds and images, we propose a novel Self-supervised Cross-modal Feature Interaction (SCFI) module. In detail, we first use paired 2D-3D proposals, predicted by the detector, to extract regional features in their respective domain. After that, a similarity loss will be exerted between paired regional features in 2D and 3D space. By interacting with cross-modal features at the instance level, SCFI strengthens the ability to perceive semantic-related information in CAFA. 

Additionally, inspired by multi-task learning, we devise a 2D-3D detection joint training paradigm to regularize the optimization of image branch. Such a training scheme prevents the overfitting issue of the image backbone and further enhances the performance of 3D detectors.

The main contributions of this work are three-fold:
\begin{itemize}
	\item We propose a learnable multi-modal feature fusion framework, called AutoAlign, which enhances the fusion process at both pixel level and instance level. 
	\item We present a joint training paradigm for 2D-3D detection to regularize the features extracted from the image branch and improve the detection accuracy. 
	\item Through extensive experiments, we validate the effectiveness of the proposed AutoAlign on various 3D detectors and achieve competitive performance on both KITTI and nuScenes datasets.   
\end{itemize}
\section{Related Work}
\subsection{3D Object Detection with Single Modality}

3D object detection is often conducted through a single modality of either RGB camera or LiDAR sensor. Camera-based 3D methods take the image as input and output the localization of objects in the space. Since monocular cameras cannot provide depth information, these models need to estimate the depth themselves~\cite{chen2016monocular}. 
For example,~\cite{mousavian20173d} first predicts the 2D bounding boxes and then estimates the depth of objects to unfold 2D boxes into 3D.
 However, monocular 3D detection often fails at predicting depth information. Therefore, stereo images are utilized to generate dense point clouds for 3D detection~\cite{you2019pseudo,li2019stereo}. 
The most widely used sensors for 3D detection are LIDARs, which can be categorized into three categories: voxel, point, and view. Voxel-based techniques discretize points into voxels and aggregate points into them to extract features~\cite{zhou2018voxelnet}. Different from voxel-based approaches, \cite{shi2019pointrcnn,yin2021center} directly process features at the point level, which maintains the original geometrical information provided by raw points, but they are generally computationally expensive. Extracting features from each view is also a popular stream in 3D detection, where points are compressed into bird-eye view~\cite{lang2019pointpillars} or range view~\cite{fan2021rangedet} for instance prediction. 

\subsection{3D Object Detection with Multi-modalities}

Recently, multi-modal fusion for object detection attracts numerous attentions. For example, \cite{qi2018frustum} predicts boxes in 2D domain and further refines them in the 3D space. \cite{ku2018joint} and~\cite{chen2017multi} attempt to perform RoI-wise fusion. In order to get more smooth BEV maps, \cite{yoo20203d} proposes to learn an auto-calibrated projection for different modalities. However, it suffers from the problem of feature blurring. Other methods~\cite{sindagi2019mvx,liang2018deep} fuse features in a point-wise manner. For instance, \cite{vora2020pointpainting} paints the 2D semantic predictions on 3D points using camera projection matrix and then performs 3D object detection. \cite{huang2020epnet} designs a novel $L_1$-Fusion module for fine-grained fusion. 

%
\begin{figure*}[!hbtp]
	\includegraphics[width=1.0\textwidth]{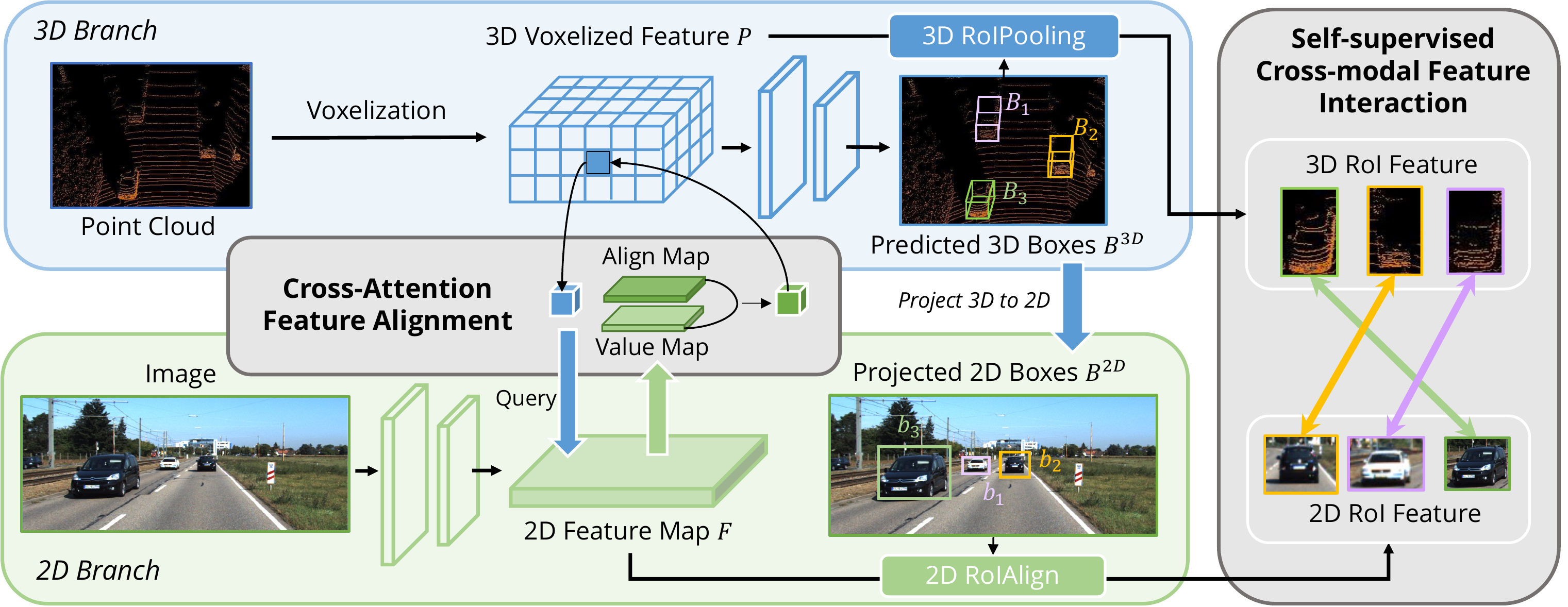}
	\caption{The framework of AutoAlign. It consists of two core components: CAFA (Sec. \ref{sec:method_cafa}) performs feature aggregation on the image plane to extract fine-grained pixel-level information for each voxel feature and SCFI (Sec. \ref{sec:method_scfi}) conducts cross-modal self-supervised supervision, exerting instance-level guidance to strengthen the semantic consistency in the CAFA module. }
	\label{fig:framework}
	\vskip -0.5em
\end{figure*}

\section{Method}

In this section, we describe the proposed AutoAlign in detail. An overview of our approach is presented in Figure \ref{fig:framework}.

\subsection{Pixel-level Feature Aggregation}
\label{sec:method_cafa}
Previous work mainly utilizes the camera projection matrix to align image and point features in a deterministic manner. This kind of approaches are effective but may bring in two potential problems: 1) the point cannot get a broader view on image data and 2) only positional consistency is maintained while ignoring semantic correlation.
Therefore, we devise the Cross-Attention Feature Alignment (CAFA) module to adaptively align the features between non-homogenous representations. Instead of adopting a one-to-one matching pattern, the CAFA module enables each voxel to perceive the whole image and dynamically attend pixel-wise 2D features based on learnable alignment maps. 

As shown in Figure \ref{fig:framework}, our method uses ResNet-50 as the backbone to extract global feature maps from given images. As a result, an input image with a size of $H \times W$ will produce the feature map with the spatial dimension of $H / 32 \times W / 32$. The feature map extracted from the image backbone is denoted as $\mathbf{Z} \in \mathbb{R}^{h\times w \times c}$, where $h, w, c$ are the height, width, and channel of the global feature map, respectively. An $1 \times 1$ convolution is added to reduce the feature dimension, creating a new feature map $\mathbf{F} \in \mathbb{R}^{h \times w \times d}$. After that, we flatten the spatial dimensions of $F$ into one dimension, resulting in a $hw \times d$ feature vector. In our cross-attention mechanism, given the feature map $F=\{f_1,f_2,...,f_{hw}\}$ ($f_i$ indicates the image feature of the $i^{th}$ spatial position) and voxel features $P=\{p_1,p_2,...,p_{J}\}$ ($p_j$ indicates each non-empty voxel feature) extracted from raw point clouds, keys and values are generated from $F$ and queries are produced by $P$. Formally,
\begin{equation}
	\mathbf{Q}_j = p_j \mathbf{W}^Q, ~~\mathbf{K}_i = f_i \mathbf{W}^K,~~ \mathbf{V}_i = f_i \mathbf{W}^V,
\end{equation}
where $\mathbf{W}^Q \in \mathbb{R}^{d\times d_k}, \mathbf{W}^K \in \mathbb{R}^{d\times d_k}$, and $\mathbf{W}^V \in \mathbb{R}^{d\times d_v}$ are linear projections. For the $j^{th}$ query $\mathbf{Q}_j$, the attention weights are calculated based on the dot-product similarity between the cross-modal query and the key:
\begin{equation}
	s_{i, j}=\frac{\exp \left(\beta_{i, j}\right)}{\sum_{j=1}^{hw} \exp \left(\beta_{i, j}\right)},
	\beta_{i, j}=\frac{{\mathbf{Q}}_{j} {\mathbf{K}}_{i}^{T}}{\sqrt{d_{k}}},
\end{equation}
where $\sqrt{d_{k}}$ is a scaling factor.
The output of the cross-attention mechanism is defined as the weighted sum over all values according to the attention weights:
\begin{equation}\label{eq:attention}
\hat{f}^{att}_{i}= \operatorname{Att}\left({\mathbf{Q}}_{i},\mathbf{K}, \mathbf{V} \right) =\sum_{j=1}^{hw} s_{i, j} {\mathbf{V}}_{j}.
\end{equation}
The normalized attention weight $s_{i, j}$ models the interests between different spatial pixels $f_{i}$ and voxel $p_{j}$, which is the \textit{align map} shown in Figure \ref{fig:framework}.
The weighted sum of the values can aggregate fine-grained
spatial pixels to update $p_{j}$, which enriches the point features with 2D information in a global view manner. Like the transformer architecture, we use the feed-forward network to produce the final RGB-aware point features as:
\begin{equation}\label{eq:FFN}
{\mathbf{F}}^{att} =\operatorname{FFN}({\mathbf{\hat{F}}}^{att}),
\end{equation}
where $\operatorname{FFN}(\cdot)$ is a simple neural network
using one fully-connected (FC) layer~\cite{vaswani2017attention}. 

\subsection{Instance-level Feature Interaction}
\label{sec:method_scfi}

The CAFA is a fine-grained paradigm in aggregating image features. However, it fails at capturing instance-level information. On the contrary, RoI-wise feature fusion maintains the integrity of the object while suffers from its coarse feature aggregation and the absence of 2D information during the proposal generation phase. 

 To bridge the gap between pixel-level and instance-level fusions, we introduce the Self-supervised Cross-modal Feature Interaction (SCFI) module to guide the learning of the CAFA. It directly utilizes the final predictions of 3D detector as proposals, which leverages both image and point features for accurate proposal generation. Moreover, instead of concatenating cross-modal features together for further box refinement, we conduct the similarity constraint between paired cross-modal features, as a manner of instance-level guidance for feature alignment.  


\begin{figure}[hbtp]
\centering
	\includegraphics[width=.35\textwidth]{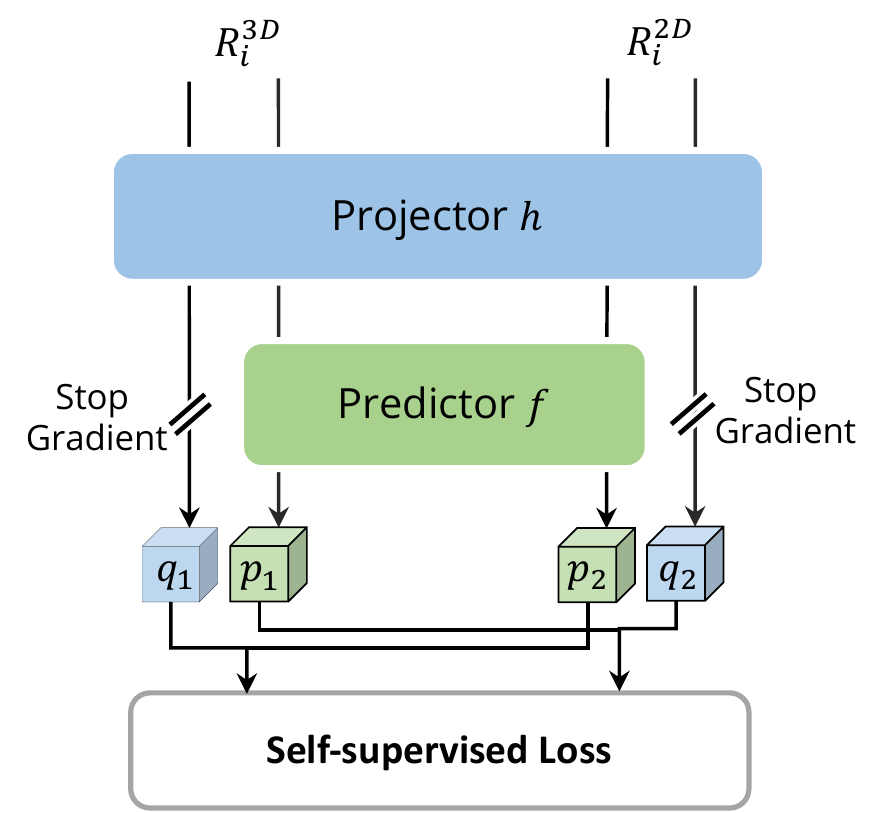}
	\caption{The architecture of Self-supervised Feature Interaction. Non-homogenous RoI features from images and points are both processed by MLP head to generate cross-modal representation for feature interaction.}
	\label{fig:cl}
 	\vskip-0.5em
\end{figure}

Given 2D feature map $\mathbf{F}$ and the corresponding 3D voxelized features $\mathbf{P}$, we randomly sample $N$ regional 3D detection boxes, denoted as $\mathbf{B}^{3D} = (B^{3D}_1,B^{3D}_2,...,B^{3D}_N)$, and then project them into the 2D plane using the camera projection matrix, resulting in a set of paired 2D boxes $\mathbf{B}^{2D} = (B^{2D}_1,B^{2D}_2,...,B^{2D}_N)$. Once obtaining the paired boxes, we adopt \texttt{2DRoIAlign}~\cite{he2017mask} and \texttt{3DRoIPooling}~\cite{shi2020points} in both 2D and 3D feature spaces to obtain respective RoI features $\mathbf{R}^{3D}$ and $\mathbf{R}^{2D}$, where each $R_{i}^{3D}$ and $R_{i}^{2D}$ are given by:
\begin{equation}
	\begin{aligned}
	R_{i}^{3D} = \texttt{3DRoIPooling}(\mathbf{P}, B^{3D}_i),\\
	R_{i}^{2D} = \texttt{2DRoIAlign}(\mathbf{F}, B^{2D}_i).
	\end{aligned}
\end{equation}

For each paired 2D and 3D RoI features, we perform self-supervised cross-modal feature interaction
on C5 from the image branch and the feature after voxelization from the point branch. Both of them are fed into a projection head $h$, transforming the output of one modality to match another modality. Similar to \cite{chen2021exploring}, a prediction head $f$ with two fully connected layers is introduced. Denoting the two output vectors as $p_1 = f(h(R^{3D}))$ and $q_2 = h(R^{2D})$, we minimize the feature distance $\mathcal{D}(p_1, q_2)$ with negative cosine similarity loss, as illustrated in Figure \ref{fig:cl}. To pull the two modal representations closer, we define a symmetry loss as:
\begin{equation}
		\mathcal{L}_{SCFI} = \frac{1}{2}\mathcal{D}(p_1, q_2) + \frac{1}{2}\mathcal{D}(p_2, q_1).
\end{equation}

Besides, the stop-gradient strategy is also adopted for the branch without prediction heads, which can be represented by $D(m_1, \texttt{stopgrad}(v_1))$. Hence, the interaction loss is implemented as:
\begin{equation}
	\mathcal{L}_{SCFI} = \frac{1}{2}D(p_1, \texttt{stopgrad}(q_2)) + \frac{1}{2}D(p_2, \texttt{stopgrad}(q_1)).
\end{equation}


\subsection{Joint Training for 2D-3D Detection}

Despite the effectiveness of multi-task learning, there is little work discussing the joint-detection for both image and point domains. In most previous methods, the image backbone is directly initialized with the pre-trained weights from other external datasets. During the training phase, the only supervision is the 3D detection loss, propagated from the point branch. Considering the large number of parameters in the image backbone, the 2D branch is more likely to get overfitting with implicit supervision. To regularize the representation extracted from the image, we extend the image branch into Faster R-CNN and supervise it with 2D detection loss, where the total loss $\mathcal{L}$ is designed as:
	\begin{gather}
		\mathcal{L} = \mathcal{L}_{3D} + \mathcal{L}_{2D} + \mathcal{L}_{SCFI},\\
		\mathcal{L}_{3D} = \mathcal{L}^{cls}_{3D} + \mathcal{L}^{reg}_{3D},\\
		\mathcal{L}_{2D} = \mathcal{L}^{cls}_{rpn} + \mathcal{L}^{reg}_{rpn} + \mathcal{L}^{cls}_{rcnn} + \mathcal{L}^{reg}_{rcnn}.
	\end{gather}


\section{Experiments}



\subsection{Implementation Details}

To validate the effectiveness of our AutoAlign, we select PointPillar~\cite{lang2019pointpillars}, SECOND~\cite{yan2018second}, and CenterPoint~\cite{yin2021center} as representative methods for our experiments. For the image branch, Faster R-CNN~\cite{ren2015faster} with ResNet50 is adopted as the 2D detector. The hidden units of cross-attention alignment module are set to 128 and the output sizes of \texttt{2DRoIAlign} and \texttt{3DRoIPooling} are both set to 4.
The MLP units of the projector and predictor of the self-supervised cross-modal module are 2048 and the hidden unit number is 512. Our 2D-3D joint training framework is optimized in an end-to-end manner with hybrid optimizers where the 3D branch is optimized with AdamW and the 2D branch is optimized with SGD. We use MMDetection3D~\cite{mmdet3d2020} as our codebase, and apply the default settings if not specified. 


\begin{table*}[!h]
	\centering
		\begin{threeparttable}
		\setlength{\tabcolsep}{1.5mm}
			\begin{tabular}{c | c | c c c | c c c | c c c | c c c }
				\toprule
				\multirow{2}*{Method} & \multirow{2}*{AutoAlign}  &  \multicolumn{3}{c|}{Car $\text{AP}_{3D}$ (\%)} & \multicolumn{3}{c|}{Pedestrian $\text{AP}_{3D}$ (\%)} & \multicolumn{3}{c|}{Cyclist $\text{AP}_{3D}$ (\%)} & \multicolumn{3}{c}{Overall $\text{AP}_{3D}$ (\%)}\\
				     & & Easy & Mod. & Hard & Easy & Mod. & Hard & Easy & Mod. & Hard & Easy & Mod. & Hard\\
				\midrule
				\multirow{2}*{PointPillar} & & 85.89 & 73.88 & 67.97 & 50.17 & 45.10 & 41.09 & 78.66 & 59.51 & 56.01 & 71.57 & 59.50 & 55.02\\
				 & \checkmark & \textbf{87.13} & \textbf{75.48} & \textbf{69.87} & \textbf{54.87} & \textbf{48.53} & \textbf{44.61} & \textbf{82.25} & \textbf{63.40} & \textbf{58.89} & \textbf{74.75} & \textbf{62.47} & \textbf{57.79}\\
				\midrule
				\multirow{2}*{SECOND} & & 87.80 & 77.47 & 74.68 & 64.73 & 59.08 & 52.84 & 83.56 & 67.42 & 62.97 & 78.70 & 67.99 & 63.50\\
					& \checkmark & \textbf{88.16} & \textbf{78.01} & \textbf{74.90} & \textbf{69.67} & \textbf{62.03} & \textbf{58.59} & \textbf{86.04} & \textbf{70.89} & \textbf{65.83} & \textbf{81.29} & \textbf{70.31} & \textbf{66.44}\\
				\bottomrule
			\end{tabular}
		\end{threeparttable}
		\vskip -0.5em
		\caption{$\text{AP}_{3D}$ performance on different 3D object detectors w/o and w/ AutoAlign on KITTI validation set.}
	\label{tab:kitti_3d}
	\vskip -0.5em
\end{table*}

\subsection{Results on KITTI dataset}
\label{sec:res_on_kitti}

In this section, we evaluate our framework on the KITTI dataset and report the average precision (AP$_{40}$). We implement AutoAlign on two representative 3D object detectors: PointPillar (pillar-based) and SECOND (voxel-based). 
The 3D mAP performance is reported in Table \ref{tab:kitti_3d}. 
Overall, our AutoAlign significantly improves PointPillar and SECOND by 3.0 and 2.3 mAP under 3D moderate evaluation protocol, which validates the effectiveness of the proposed method. When observing the results in detail, we find that the APs of Pedestrian and Cyclist are promoted most (3.0 and 3.5 mAP on $\text{AP}_{3D}$ moderate, respectively). We infer the reason that cars often hold more points, while objects like pedestrians and cyclists are mostly short of reflection, which makes them harder to be detected in the 3D space. Therefore, AutoAlign benefits from the RGB data that are naturally dense and rich in semantic and texture information.


\subsection{Results on NuScenes dataset}

We also conduct experiments on the much larger nuScenes dataset with current state-of-the-art 3D detector CenterPoint to further validate the effectiveness of AutoAlign. As shown in Table~\ref{tab:nus_val}, AutoAlign achieves 66.6 mAP and 71.1 NDS on the nuScenes validation set, outperforming the strong CenterPoint baseline by 7.0 mAP and 6.5 NDS. It also surpasses the recently developed multi-modal 3D detector MVP~\cite{yin2021multimodal} by 1.1 NDS under the same single-stage settings. Besides, due to its simplicity and joint-training paradigm, it does not require any sophisticated virtual point generation or image feature pre-fetching, which is much more suitable for real-world applications. We also report detailed results on each object category as well as the performance on the test leaderboard in the supplementary material.
\begin{table*}[!hbpt]
	\centering
		\begin{threeparttable}
			\begin{tabular}{c c c | c c c | c c c }
				\toprule
				\multirow{2}*{CAFA (pixel-level)} & \multirow{2}*{SCFI (instance-level)} & \multirow{2}*{2D Joint Training} & \multicolumn{3}{c|}{$\text{AP}_{3D}$ (\%)} & \multicolumn{3}{c}{$\text{AP}_{BEV}$ (\%)}\\
				 &  &  & Easy & Mod. & Hard & Easy & Mod. & Hard\\
				\midrule
					& & & 78.70 & 67.99 & 63.50 & 81.05 & 74.42 & 70.86\\
				\checkmark & & & 79.64 & 68.54 & 64.24 & 81.35 & 75.13 & 71.34\\
				\checkmark & \checkmark & & 80.63 & 69.67 & 65.49 & 82.13 & 76.04 & 72.53\\
				\checkmark & \checkmark & \checkmark & \textbf{81.29} & \textbf{70.31} & \textbf{66.44} & \textbf{83.68} & \textbf{77.71} & \textbf{73.72}\\
				\bottomrule
			\end{tabular}
		\end{threeparttable}
		\vskip -0.5em
		\caption{Effect of each component in our AutoAlign. Results are reported on KITTI validation set with SECOND.}
	\label{tab:kitti_ablation}
	\vskip -0.8em
\end{table*}

\begin{table}[!h]
	\centering
		\begin{tabular}{c |c c |c}
			\toprule
			Method & mAP & NDS & Reference\\
			\midrule
			\midrule
			PointPainting  & 45.6 & 54.6 & CVPR~\citeyear{vora2020pointpainting} \\
			3D-CVF & 42.1 & 49.8 & ECCV~\citeyear{yoo20203d}\\
			CenterPoint & 56.4 & 64.8 & CVPR~\citeyear{yin2021center} \\
			ObjectDGCNN  & 58.6 & 66.0 & NeurIPS~\citeyear{wang2021object} \\
			MVP & 66.0 & 70.0 & NeurIPS~\citeyear{yin2021multimodal}\\
			\midrule
			CenterPoint* & 59.6 & 66.6 & CVPR~\citeyear{yin2021center} \\
			CenterPoint + AutoAlign & \textbf{66.6} & \textbf{71.1} & -\\
			\bottomrule
		\end{tabular}
	\vskip -0.25em
	\caption{mAP and NDS performance on nuScenes dataset. The models are trained on nuScenes train subset and evaluated on nuScenes validation subset. * indicates our re-implementation.}
	\label{tab:nus_val}
	\vskip -0.5em
\end{table}

%

\subsection{Ablation Studies}

To understand how each module in AutoAlign promotes the detection accuracy, we test each component on the baseline detector SECOND and report its AP performance on the KITTI validation dataset in Table \ref{tab:kitti_ablation}. 

When cross-attention feature alignment is applied, the accuracy is raised by 0.5 mAP and the improvements are found on objects of all difficulty levels. This result validates the importance of preserving the high resolution of image information when aggregating cross-modal features.  

Then, we add the SCFI module which brings in a 1.2 mAP enhancement, namely the overall moderate $\text{AP}_{3D}$ improves from 68.5 to 69.7, suggesting that feature interaction plays a pivotal role in our fusion framework. It exerts instance-level supervision on the feature alignment, which hints at how to aggregate semantically paired features across non-homogenous representations. 

When 2D joint training is added, the accuracy is boosted by another 0.6 mAP and the $\text{AP}_{hard}$ is raised by 1.0 mAP. Such a large improvement benefits from two aspects: 1) the joint training paradigm regularizes the optimization of image backbone and 2) joint optimization reduces the training gap between 2D and 3D models and maintains the feature consistency during the cross-modal feature fusion process. 

\subsection{Discussions}

In this section, we delve into AutoAlign framework to study how the detection accuracy is achieved and gain a deeper understanding of the underlying mechanisms. For all experiments, we employ SECOND with the same settings in Section \ref{sec:res_on_kitti}. 

\begin{figure*}[!hbtp]
\centering
	\includegraphics[width=0.9\textwidth]{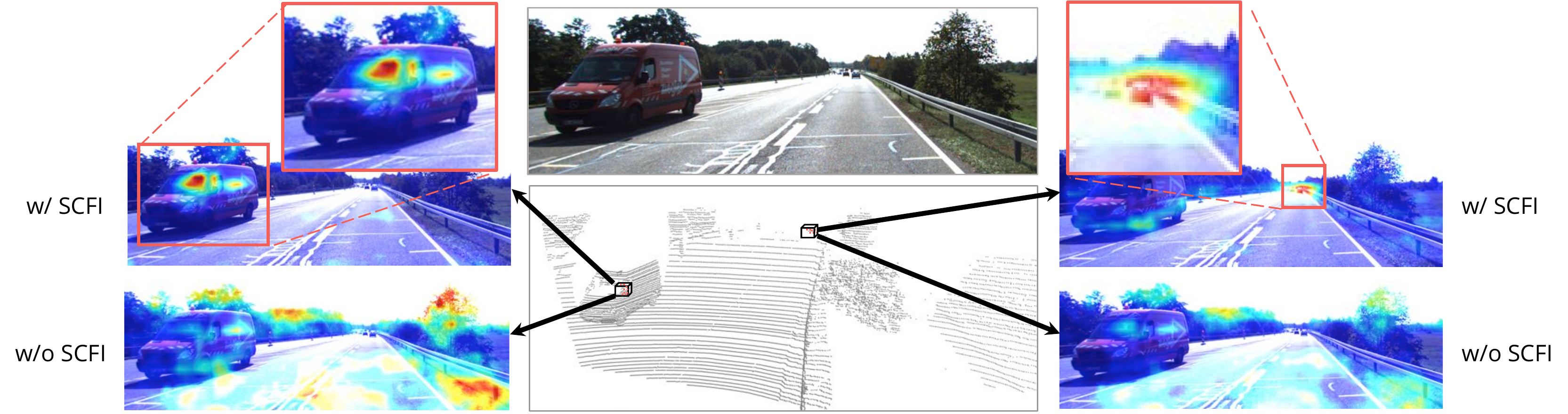}
	\caption{Visualization of alignment maps generated by CAFA module from two randomly selected point voxels. To validate the effectiveness of SCFI module, we also visualize the alignment map without SCFI module. SCFI regularizes CAFA with instance-level semantic supervision, resulting in a positionally and semantically meaningful alignment map.}
	\label{fig:query_vis}
	\vskip -1em
\end{figure*}

	\subsubsection{Investigating the Best Cross-modal Query Strategy.} 
	In this part, we compare various strategies for cross-modal feature query. Firstly, we choose the widely adopted fusion strategy, \textit{i.e.}, projecting points to the image plane through the camera projection matrix and utilizing point-wise bilinear-interpolation to obtain the aligned 2D image. Secondly, we test the non-local block proposed in~\cite{wang2018non}, where all image features are taken into account but only interested positions with high attention scores will be aggregated for cross-modal fusion. Finally, we adopt a more general form, which is similar to the self-attention module in~\cite{vaswani2017attention}, but we extend it from the same modalities to non-homogenous representations. Following the common design of self-attention, we explore the performance of single-head cross-attention module and the multi-head one. The detailed results are listed in Table \ref{tab:ablation_query}. When using point-based projection, the improvement is limited, since the points are unable to get continuous image features. However, when replacing the point-based projection with non-local block, the performance is still unsatisfying. The possible reason may lie in the FC layers that lead to the overfitting issue. Compared to vanilla non-local block, the performance of cross-attention is more competitive, probably due to the adoption of dropout strategy and feature normalization. Considering the computational cost and efficiency, we finally take single-head cross-attention as our query strategy.
	
	\begin{table}[!h]
		\centering
		\begin{threeparttable}
			\begin{tabular}{c | c c c}
				\toprule
				\multirow{2}*{Query Strategy}  &    & $\text{AP}_{3D}$ (\%) & \\
				   & Easy & Mod. & Hard\\
				 \midrule
				 Point-based Proj & 80.24 & 69.40 & 65.65\\
				 Non-Local & 80.01 & 69.13 & 65.34\\
				 Multi-head Cross-attention & 81.04 & 70.25 & \textbf{66.49}\\
				 Single-head Cross-attention & \textbf{81.29} & \textbf{70.31} & 66.44\\
				\bottomrule
			\end{tabular}
		\end{threeparttable}
		\vskip -0.25em
		\caption{$\text{AP}_{3D}$ performance with various query strategies for cross-modal feature alignment.}
	\label{tab:ablation_query}
	\vskip -1em
\end{table}
	
	\subsubsection{Seeking the Suitable Feature Source for Self-supervised Feature Interaction.}
	
	Feature interaction is a core component since it intensifies the semantic consistency of the CAFA module with instance-level guidance. Hence, how to select suitable feature source for self-supervised learning is non-trivial. After carefully examining the selection of point feature and image feature sources, we take the image features directly from ResNet backbone (\textit{i.e.}, C5) and after FPN (\textit{i.e.}, P5) as candidates. For the point branch, we select the features before point backbone and after backbone. As shown in Table \ref{tab:feat_source}, using C5 as the image feature is better than P5. We infer the reason that P5 is directly for 2D detection and therefore limits the generalization ability for cross-modal feature fusion, while C5 is more flexible for both 2D detection and non-homogenous self-supervised learning. When choosing point features after backbone, we observe a quick convergence of the similarity loss, but the result is unsatisfying. This may stem from the too much flexibility of the 3D branch, \textit{i.e.}, the point backbone provides possibility of complex transformation for point features, which eases the optimization of loss but weakens the instance-level guidance of semantic consistency by our proposed self-supervised feature interaction. On the contrary, although using the features before backbone slows down the convergence, the model is implicitly supervised by mutual interaction and gradually learns how to align cross-modal features in the CAFA module.
    
    \begin{table}[!h]
	\centering
		\begin{threeparttable}
			\begin{tabular}{c | c | c c c}
				\toprule
				\multirow{2}*{Img Feat} & \multirow{2}*{Pts Feat} & \multicolumn{3}{c}{$\text{AP}_{3D}$(\%)}\\
				 & & Easy & Mod. & Hard\\
				 \midrule
				 P5 & after backbone & 78.42 & 67.49 & 63.98\\
				 C5 & after backbone & 79.14 & 68.28 & 64.73\\
				 P5 & before backbone & \textbf{81.53} & 70.06 & 66.11\\
				 C5 & before backbone & 81.29 & \textbf{70.31} & \textbf{66.44}\\
			\bottomrule
			\end{tabular}
		\end{threeparttable}
	\vskip -0.5em
	\caption{$\text{AP}_{3D}$ performance with different feature sources from image/points for cross-modal feature interaction.}
	\label{tab:feat_source}
	\vskip -1em
\end{table}

	\subsubsection{Optimal Loss for Self-supervised Cross-modal Learning.}

	Since most self-supervised learning approaches are based on homogenous representations, exploring the optimal self-supervised loss for cross-modality is necessary. We compare four different prototypes and report the results in Table \ref{tab:ablation_loss}. We adopt the classical version of contrastive loss, where both positive and negative pairs are considered. Note that features located at the same position in 3D space and 2D plane are considered as positive pairs while the rest ones are negative pairs. The selections of NCE loss and its variant InfoNCE do not provide remarkable enhancement. However, when utilizing positive pairs only for feature interaction, we observe significant improvements. We infer the reason that the points hold less identity information compared to images. When supervising two similar instances with negative pair loss, it may deteriorate the 3D feature representation if the shapes of the instances are similar to each other. 
Therefore, we choose the negative cosine similarity loss with positive pairs for our feature interaction module.  
\begin{table}[!h]
	\centering
		\begin{threeparttable}
			\begin{tabular}{c | c | c c c}
				\toprule
				\multirow{2}*{Loss} & \multirow{2}*{Pair}  &    \multicolumn{3}{c}{$\text{AP}_{3D}$(\%)}\\
				 &  & Easy & Mod. & Hard\\
				\midrule
				NCE Loss & Pos + Neg & 79.14 & 68.65 & 65.54\\
				InfoNCE Loss & Pos + Neg & 79.63 & 68.57 & 65.21\\
				CE Loss & Pos & 80.83 & 69.73 & 66.01\\
				NCS Loss & Pos & \textbf{81.29} & \textbf{70.31} & \textbf{66.44}\\ 
				\bottomrule
			\end{tabular}
		\end{threeparttable}
		\vspace{-0.5em}
		\caption{$\text{AP}_{3D}$ performance with different loss settings for self-supervised cross-modal learning. (NCS: negative cosine similarity)}
	\label{tab:ablation_loss}
	\vspace{-1.5em}
\end{table}

\subsection{Visualization and Analysis}

In addition to the detection results on various datasets, we also provide a more straightforward visualization on the learned alignment map, given different 3D query voxel features, as shown in Figure \ref{fig:query_vis}. To better illustrate the mutual influence between the CAFA and SCFI modules, we compare the query attention map with and without SCFI. It can be clearly concluded that CAFA fails to yield a meaningful alignment map on 2D images without feature interaction. Conversely, when armed with SCFI, the CAFA module can successfully provide a positionally and semantically reasonable feature alignment map. 
\vskip -1em

\section{Conclusion}

In this work, we develop AutoAlign, a learnable multi-modal feature fusion method for 3D object detection. The proposed Cross-Attention Feature Alignment module enables each voxel feature to aggregate image information in a fine-grained manner. Furthermore, a novel Self-supervised Cross-modal Feature Interaction module is designed to enhance the semantic consistency during assignment for the CAFA module. Comprehensive experimental results have demonstrated that AutoAlign significantly improves various 3D detectors on the KITTI and nuScenes datasets. We hope our work could provide a new perspective in multi-modal feature fusion for autonomous driving.

\newpage
\bibliographystyle{named}
\bibliography{ijcai19}
\end{document}